\documentclass[10pt, a4paper]{article}

\usepackage{lrec2026}

\usepackage{tikz}
\usetikzlibrary{matrix, positioning, backgrounds, calc}

\usepackage{times}
\usepackage{latexsym}
\usepackage{subcaption}
\usepackage[utf8]{inputenc}
\usepackage[T1]{fontenc}
\usepackage{CJKutf8}
\usepackage{microtype}
\usepackage{inconsolata}
\usepackage{graphicx}
\usepackage[capitalise]{cleveref}
\usepackage{booktabs}
\newcommand*{\jpn}[1]{\begin{CJK}{UTF8}{ipxg}#1\end{CJK}}
\newcommand{\bis}{\textbf{BIS Reasoning 1.0}}
\newcommand{\llm}[1]{\texttt{#1}}
\newcommand{\TODO}[1]{\textcolor{blue}{(TODO) #1}}

\title{BIS Reasoning 1.0: The First Large-Scale Japanese Benchmark for Belief-Inconsistent Syllogistic Reasoning}

\name{Ha-Thanh Nguyen\thanks{Corresponding email: nguyenhathanh@nii.ac.jp}, Hideyuki Tachibana, Chaoran Liu, Qianying Liu,\\
{\large\bfseries  Su Myat Noe, Koichi Takeda, Sadao Kurohashi}
}

\address{Research and Development Center for Large Language Models, NII, Tokyo, Japan}

\abstract{
We present \bis{}, the first large-scale Japanese dataset of syllogistic reasoning problems explicitly designed to evaluate belief-inconsistent reasoning in large language models (LLMs). Unlike prior resources such as NeuBAROCO and JFLD, which emphasize general or belief-aligned logic, \bis{} systematically introduces logically valid yet belief-inconsistent syllogisms to expose belief bias—the tendency to accept believable conclusions irrespective of validity. 
We benchmark a representative suite of cutting-edge models—including OpenAI GPT-5 variants, GPT-4o, Qwen, and prominent Japanese LLMs—under a uniform, zero-shot protocol. Reasoning-centric models achieve near-perfect accuracy on \bis{} (e.g., \llm{Qwen3-32B} $\approx$99\% and \llm{GPT-5-mini} up to $\approx$99.7\%), while \llm{GPT-4o} attains around 80\%. Earlier Japanese-specialized models underperform, often well below 60\%, whereas the latest \llm{llm-jp-3.1-13b-instruct4} markedly improves to the mid-80\% range. These results indicate that robustness to belief-inconsistent inputs is driven more by explicit reasoning optimization than by language specialization or scale alone.
Our analysis further shows that even top-tier systems falter when logical validity conflicts with intuitive or factual beliefs, and that performance is sensitive to prompt design and inference-time reasoning effort. We discuss implications for safety-critical domains—law, healthcare, and scientific literature—where strict logical fidelity must override intuitive belief to ensure reliability.
\\ \newline
}

\begin{document}

% Generate title and abstract on the first page.  The LREC macro
% \maketitleabstract combines \maketitle and the abstract.
\maketitleabstract

% -------------------- Main text begins here --------------------

\section{Introduction}

Large language models (LLMs) have demonstrated remarkable performance on various natural language tasks, yet ensuring reliable logical reasoning remains an open challenge \cite{morishita-etal-2024-jfld,nguyen2023negation}. This issue is particularly critical in high-stakes domains such as law, healthcare, and scientific research, where even subtle reasoning errors can lead to severe consequences \cite{yan-etal-2025-llm}. Alarmingly, recent studies show that the persuasive fluency of models like GPT-4 can deceive users into trusting incorrect conclusions \cite{bajpai2024can}. Hence, rigorously evaluating and enhancing LLM reasoning accuracy is crucial before deploying them in domains demanding strict logical rigor.

A major concern in current LLM research is their susceptibility to human-like cognitive biases, notably the belief bias -- accepting conclusions aligned with prior beliefs regardless of logical validity. This bias poses significant risks in applications requiring impartial logic. For example, \citet{dasgupta2022language} demonstrated that LLMs frequently endorse logically invalid arguments simply because their conclusions appear believable, highlighting a critical vulnerability.

Existing benchmarks for evaluating logical reasoning in LLMs exhibit several limitations. Most influential reasoning benchmarks are predominantly English-based \cite{li-etal-2023-counterfactual,qin-etal-2019-counterfactual,frohberg-binder-2022-crass}, creating a significant evaluation gap for languages such as Japanese. Although some Japanese datasets exist, like JFLD \cite{morishita-etal-2024-jfld}, which tests formal logic isolated from real-world knowledge, and NeuBAROCO \cite{ozeki2024exploring}, which covers multiple biases but has limited belief-inconsistent content, no dedicated Japanese-language dataset explicitly targets belief-inconsistent reasoning.

\begin{figure}
\centering
\includegraphics[width=0.9\linewidth]{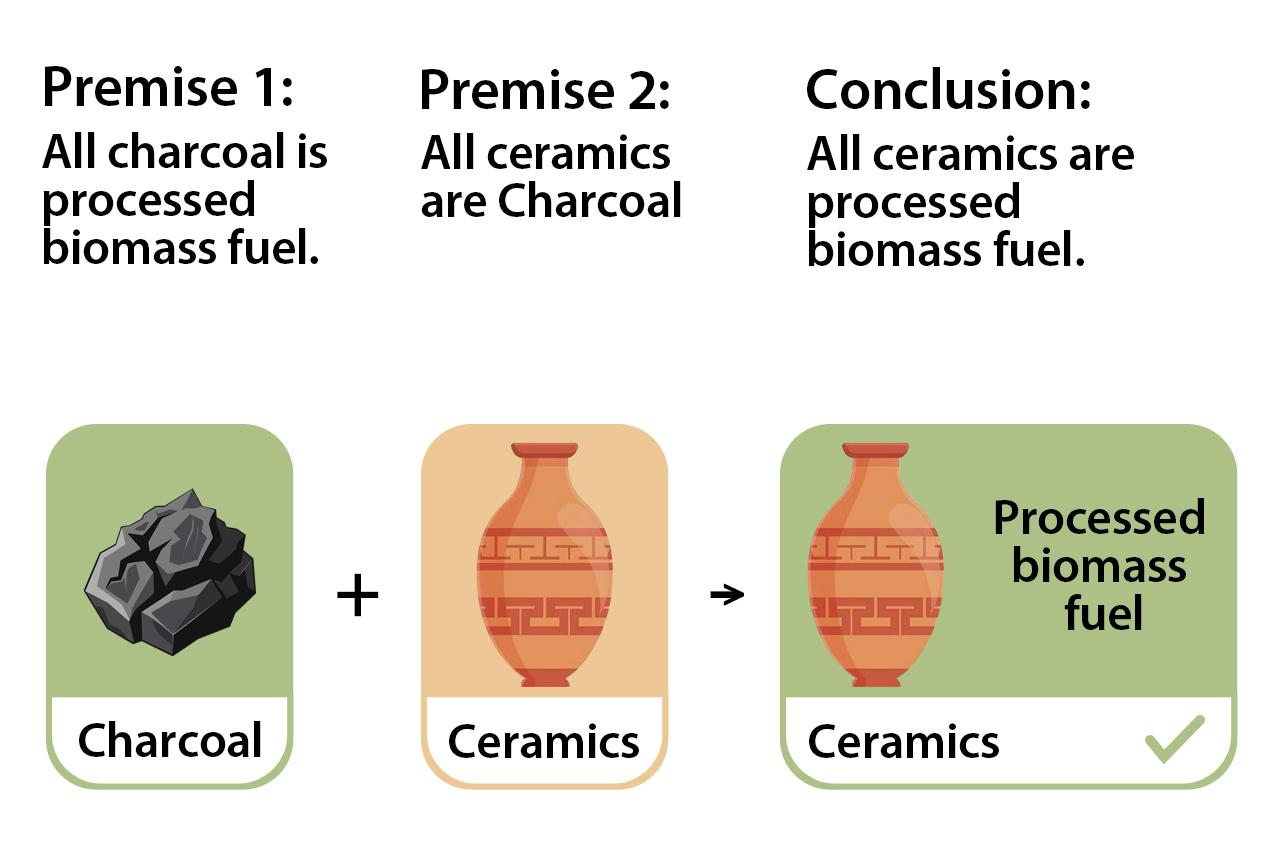}
\caption{A translated example from the BIS Dataset illustrating a belief-inconsistent syllogism: although the conclusion is logically valid, it contradicts common real-world beliefs.}
\label{fig:example}
\end{figure}

To address this gap, we introduce \bis{} -- the first Japanese dataset designed explicitly for assessing belief-inconsistent syllogistic reasoning in LLMs. BIS (Belief-Inconsistent Syllogisms) aims to evaluate whether models can uphold logical validity when correct conclusions conflict with typical beliefs or factual knowledge. \cref{fig:example} shows a representative example from the BIS Dataset, illustrating how logically valid conclusions can defy intuitive beliefs. Specifically, our contributions include:

\begin{enumerate}
    \item \textbf{\bis{} Dataset:} We present a carefully curated collection of Japanese syllogistic reasoning problems explicitly constructed to challenge LLMs with logically valid conclusions that contradict common beliefs. This dataset enables the first targeted evaluation of belief-inconsistent reasoning capabilities in Japanese LLMs.

    \item \textbf{Comprehensive Evaluation of Leading LLMs:} We benchmark state-of-the-art models --
    %including GPT-4o, GPT-4-turbo, Claude, and prominent Japanese LLMs
    including OpenAI GPT, Claude, Qwen and prominent Japanese LLMs
    -- under standardized conditions, providing the first systematic comparison of their performance on belief-inconsistent reasoning in Japanese.

    \item \textbf{Detailed Analysis of Performance and Bias:} Our analysis identifies significant performance gaps, highlighting that even advanced LLMs struggle disproportionately with belief-inconsistent problems. We quantify these biases, investigate specific syllogistic structures prone to errors, and examine how prompts and CoT (chain-of-thought) approaches affect reasoning accuracy.

    \item \textbf{Implications for Reliability in Real-World Applications:} We discuss critical implications for deploying LLMs in safety-critical domains. \bis{} reveals vulnerabilities that standard benchmarks typically overlook, providing crucial insights for improving logical consistency and objectivity in real-world scenarios like law, medicine, and scientific research.
\end{enumerate}

Overall, \bis{} contributes significantly to understanding and improving logical reasoning in Japanese-language LLMs. By explicitly evaluating belief-inconsistent reasoning, this work advances efforts toward creating reliable, bias-resistant models suitable for deployment in high-stakes environments.

\section{Related Work}

Evaluating the logical reasoning abilities of large language models (LLMs) has become a key research focus. Benchmarks like ReClor \cite{yureclor} and LogiQA \cite{liu2023logiqa} use multiple-choice logic problems derived from standardized exams to test inference beyond surface semantics. Despite advances such as chain-of-thought prompting \citep{wei2022chain,kojima2022large} and neuro-symbolic modeling, LLMs still struggle to match human performance on tasks requiring rigorous logic.

LLMs not only make logical errors but also exhibit human-like cognitive biases. One well-studied bias is belief bias -- the tendency to accept conclusions that align with prior beliefs regardless of logical validity \cite{evans1983conflict}. Studies have shown that LLMs are more accurate on belief-consistent reasoning tasks and frequently misjudge belief-inconsistent syllogisms \cite{ando-etal-2023-evaluating, ozeki2024exploring}.

Instruction tuning and RLHF can amplify these tendencies. For instance, models like GPT-4 and Claude, while highly fluent, may reinforce belief-aligned reasoning due to human preferences during fine-tuning \cite{bai2022training}. This further underscores the need for benchmarks that reveal latent cognitive biases and test reasoning under belief-conflicting conditions.

In the Japanese language, recent efforts have produced datasets for logical reasoning evaluation, yet limitations remain. JFLD \cite{morishita-etal-2024-jfld} focuses on formal deductive reasoning using artificially constructed propositions to isolate logic from world knowledge. While large in scale and diverse in structure, its use of semantically unnatural sentences and synthetic vocabulary prevents assessment of reasoning in realistic settings. JaNLI \cite{yanaka2021assessing} and JAMP \cite{sugimoto-etal-2023-jamp} explore adversarial and temporal inference respectively, but they do not test belief-inconsistent reasoning and lack syllogistic structure. NeuBAROCO \cite{ozeki2024exploring} closely aligns with our goals, exploring belief bias in syllogistic reasoning. However, it falls short as a comprehensive benchmark: the Japanese subset contains fewer than 800 examples for the NLI task and under 100 for the multiple-choice format, and it does not exclusively target belief-inconsistent reasoning. In contrast, \bis{} offers a focused and large-scale evaluation specifically designed to test logical robustness under belief-conflicting conditions.

These observations point to a significant gap: current Japanese benchmarks either employ unnatural representations, ignore belief-inconsistent logic, or lack scale and coverage of syllogistic forms. While prior work has identified belief bias in LLMs, comprehensive datasets for evaluating such bias in Japanese syllogistic reasoning remain scarce. \bis{} directly addresses this gap by providing a focused, large-scale, and naturalistic benchmark specifically designed to test how LLMs handle logically valid conclusions that conflict with intuitive beliefs.

\begin{figure}[t]
\centering
\includegraphics[width=\linewidth]{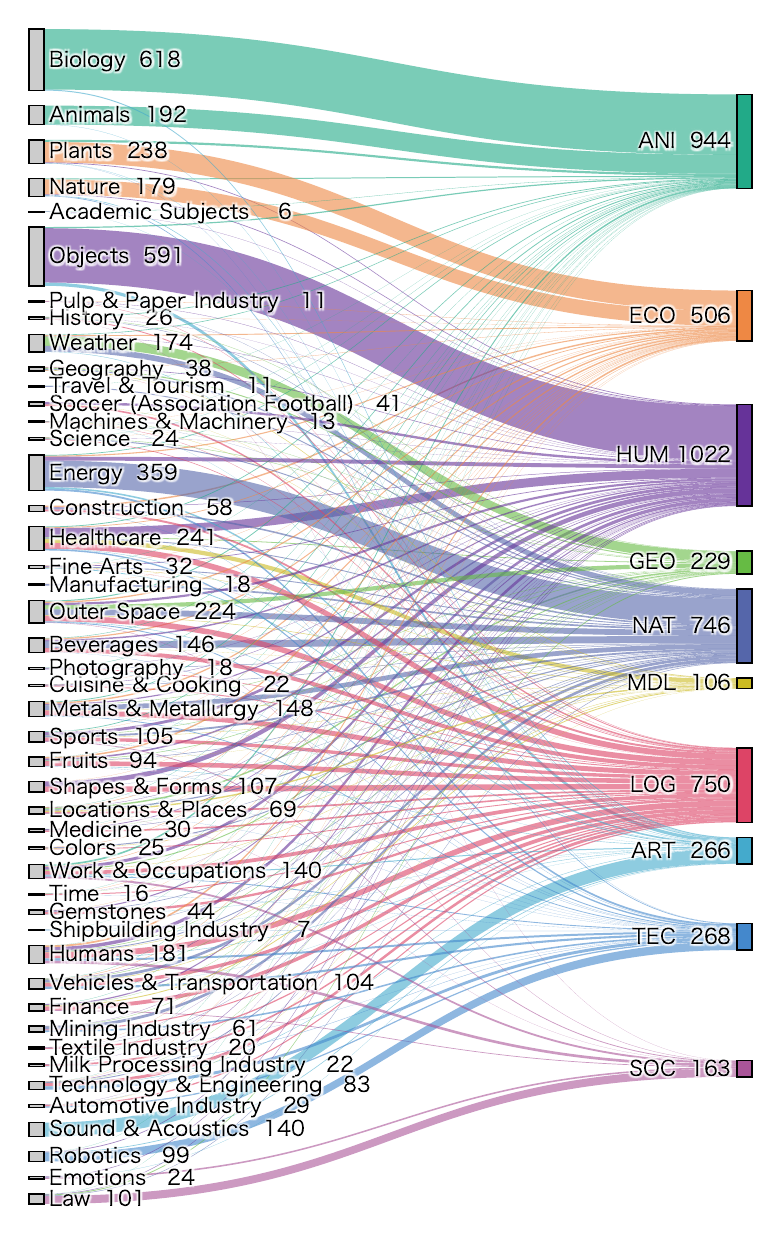}
% I tried to create another figure. If you like it, feel free to use it.
%\caption{Combined category analysis of top raw and final semantic categories in the BIS dataset. The figure illustrates the top 5 raw categories (left), top 5 final categories (right), and the heatmap shows the relationship between these raw and final categories, highlighting category intersections and distribution patterns.}
\caption{Combined category analysis of raw categories (left) and 10 final categories: animals \& living things, ecosystem, human body \& senses, geology, natural phenomena, models, logic \& structure, arts, technology and society. }
\label{fig:combined_category_analysis}
\end{figure}

\section{Dataset Construction}

We present \bis{}, a Japanese-language dataset consisting of 5,000 carefully constructed syllogistic reasoning problems. The dataset is specifically designed to test the robustness of logical inference in large language models (LLMs) under conditions of \textit{belief inconsistency}, where logically valid conclusions explicitly contradict widely held commonsense beliefs. \bis{} serves as a diagnostic benchmark for probing whether LLMs can prioritize formal logic over prior-knowledge heuristics in natural language reasoning.

The dataset was developed through a formalized specification process aimed at ensuring both logical rigor and linguistic quality. Each example comprises two premises and one conclusion that is strictly entailed by syllogistic rules, such as classic forms like ``\textit{All A are B; All C are A; Therefore, All C are B.}'' Crucially, the conclusions are deliberately chosen to conflict with general knowledge, encouraging model errors driven by belief bias. This design isolates the logical reasoning process from superficial semantic plausibility, exposing potential biases embedded in LLM training data.

To ensure linguistic fluency and naturalness, all annotators involved in dataset construction were either native Japanese speakers or individuals with advanced Japanese proficiency. Initially, the dataset covers 46 distinct semantic categories (raw categories), ranging from concrete areas like animals, food, and weather to abstract domains such as law, logic, and emotion (left side of \cref{fig:combined_category_analysis}). These detailed raw categories were subsequently consolidated into 10 broader final categories to facilitate interpretability, ensure topic balance, and support higher-level reasoning analysis (right side of \cref{fig:combined_category_analysis}). %Figure~\ref{fig:combined_category_analysis} summarizes the distribution and relationship between these raw and final categories, reflecting a concentration in cognitively rich areas valuable for probing belief-inconsistent reasoning in LLMs.

All examples underwent a two-phase quality assurance (QA) process, initially involving manual review of 10\% of examples for iterative feedback, followed by comprehensive review to ensure structural validity, language clarity, and semantic diversity. Issues identified included syntactic violations, duplicated content, ambiguous premises, and category imbalance.

\section{Experiments}
\begin{table*}
\centering
%\begin{tabular}{@{}lccp{7cm}@{}}
\begin{tabular}{@{}lcc@{}}
\toprule
%\textbf{Model} & \textbf{BISR1.0} & \textbf{NeuB} 
\textbf{Model} & \textbf{BIS Reasoning 1.0} & \textbf{NeuBAROCO} 
%& \textbf{Response Sample} 
\\
%%%%%%%%%%%%%%%%%%%%%%%%%%%%%%%%%%%%%%%%%%%%%%%%%%%%%%%%%%%%%
\midrule
\llm{GPT-5-mini} (\textit{medium} reasoning effort)
& 99.72 & 91.92 % exact match
%\scriptsize{\jpn{はい。}\checkmark}
\\
%%%%%%%%%%%%%%%%%%%%%%%%%%%%%%%%%%%%%%%%%%%%%%%%%%%%%%%%%%%%%
\llm{GPT-5-nano} (\textit{medium} reasoning effort)
& 98.84 & 91.32 % exact match
%\scriptsize{\jpn{はい。}\checkmark}
\\
%%%%%%%%%%%%%%%%%%%%%%%%%%%%%%%%%%%%%%%%%%%%%%%%%%%%%%%%%%%%%
\llm{gpt-oss-20b} 
& 98.56 & 89.52  % heuristics
%\scriptsize{analysisWe have the problem in Japanese: \jpn{以下の三段論法を考えてください} [...] Thus I'd say "\jpn{はい}". So answer: \jpn{はい}. assistantfinal\jpn{はい}\checkmark} 
\\ 
%%%%%%%%%%%%%%%%%%%%%%%%%%%%%%%%%%%%%%%%%%%%%%%%%%%%%%%%%%%%%
\llm{GPT-4o} 
& 79.54 & 94.01 % exact match
%\scriptsize{\jpn{はい。}\checkmark} 
\\
% I obtained similar results for GPT-4o (using Azure API)
%%%%%%%%%%%%%%%%%%%%%%%%%%%%%%%%%%%%%%%%%%%%%%%%%%%%%%%%%%%%%
\llm{GPT-5-nano} (\textit{minimum} reasoning effort)
& 69.22 & 73.05 % exact match
%\scriptsize{\jpn{はい。}\checkmark}
\\
%%%%%%%%%%%%%%%%%%%%%%%%%%%%%%%%%%%%%%%%%%%%%%%%%%%%%%%%%%%%%
\llm{GPT-4-turbo} 
& 59.48 & 67.66 % exact match
%\scriptsize{\jpn{はい。}\checkmark}
\\
\midrule
%%%%%%%%%%%%%%%%%%%%%%%%%%%%%%%%%%%%%%%%%%%%%%%%%%%%%%%%%%%%%
\llm{Qwen3-32B} (w/o thinking) 
& 99.58  & 94.01  % exact matching
%\scriptsize{\jpn{はい。}\checkmark} 
\\ 
%%%%%%%%%%%%%%%%%%%%%%%%%%%%%%%%%%%%%%%%%%%%%%%%%%%%%%%%%%%%%
\llm{Qwen3-32B} (w/ thinking) 
& 99.12  & 97.60  % heuristics
%\scriptsize{<think>Alright, let's tackle this syllogism problem. So, the user provided three statements in Japanese and wants to know if the conclusion logically follows from the premises. [...] So the answer should be "Yes."</think>\jpn{はい}\checkmark} 
\\ 
%%%%%%%%%%%%%%%%%%%%%%%%%%%%%%%%%%%%%%%%%%%%%%%%%%%%%%%%%%%%%
\midrule
%%%%%%%%%%%%%%%%%%%%%%%%%%%%%%%%%%%%%%%%%%%%%%%%%%%%%%%%%%%%%
\llm{llm-jp-3-13b-instruct4} 
& 84.66 & 73.65 % llm-as-a-judge
%\scriptsize{\jpn{[...] したがって、論理構造自体は正しいものの、提示された内容に関しては誤った情報を含んでいるため、その結論はその前提から直接導かれるものではありません。}\crossmark}
\\
%%%%%%%%%%%%%%%%%%%%%%%%%%%%%%%%%%%%%%%%%%%%%%%%%%%%%%%%%%%%%
\llm{stockmark-13b} 
& 55.90 & 60.48 % llm-as-a-judge
%\scriptsize{\jpn{[...] 前提1と前提2から結論を導くことは不可能である。\nなぜなら、前提1は仮定であり、すべての海底が海洋で最も深い場所であるとは証明されていないからである。\n [...]}\crossmark} 
\\
%%%%%%%%%%%%%%%%%%%%%%%%%%%%%%%%%%%%%%%%%%%%%%%%%%%%%%%%%%%%%
\llm{llm-jp-3-13b} 
& 34.78 & 36.52 % llm-as-a-judge
%\scriptsize{\jpn{[...] この三段論法では最初の二つの前提を満たすような事例がたくさんあります。そのため、この三段論法は正しくありません。\crossmark}}
\\

%%%%%%%%%%%%%%%%%%%%%%%%%%%%%%%%%%%%%%%%%%%%%%%%%%%%%%%%%%%%%
\llm{llm-jp-3-13b-instruct3} 
& 11.06 & 22.46 % llm-as-a-judge
%\scriptsize{\jpn{[...] したがって、論理構造自体は正しいものの、提示された内容に関しては誤った情報を含んでいるため、その結論はその前提から直接導かれるものではありません。}\crossmark}
\\
%%%%%%%%%%%%%%%%%%%%%%%%%%%%%%%%%%%%%%%%%%%%%%%%%%%%%%%%%%%%%
\midrule
\llm{Claude-3-sonnet-20240229} (\textit{deprecated}) & 20.34 & 78.44 \\
% This claude model is now deprecated.
%%%%%%%%%%%%%%%%%%%%%%%%%%%%%%%%%%%%%%%%%%%%%%%%%%%%%%%%%%%%%
\llm{Claude-3-opus-20240229} (\textit{deprecated}) & \phantom{0}7.18 & 61.07 \\
% Opus 20240229 is still working but will be deactivated soon.
\bottomrule
\end{tabular}
\caption{Accuracy of models on the \bis{} dataset and NeuBAROCO belief-inconsistent syllogisms. All the results are based on the `Basic' prompt in \cref{tab:prompt_descriptions}. %All the response samples are for the first entry of \bis{} dataset. 
}
\label{tab:overall_performance}
\end{table*}

\subsection{General Settings}

To evaluate how well LLMs handle logically valid yet belief-inconsistent inferences, we formulate \bis{} as a diagnostic test focusing strictly on logical judgment. Specifically, models must determine if a given conclusion logically follows from two premises, even when the conclusion contradicts intuitive beliefs.

We framed the task as a binary classification using concise, instruction-based prompts in Japanese, asking models to judge logical entailment by answering ``Yes'' or ``No.'' %``\jpn{はい}'' (\textit{Yes}) or ``\jpn{いいえ}'' (\textit{No}). 
Each prompt clearly stated two premises and one conclusion, accompanied by a brief system instruction reinforcing the model’s logical reasoning role as shown in \cref{tab:prompt_descriptions}. 
% (TODO) necessity of the below sentence
%Models were not required to provide explanations, thereby isolating pure logical inference capability from linguistic fluency or explanatory quality.
LLMs were not explicitly required nor prohibited to provide explanations.
To derive the Yes/No labels from the LLM responses, we primarily utilized heuristic pattern matching. In cases where this approach fell short due to lengthy elucidations by the LLMs, we employed the LLM-as-a-judge strategy, using the \llm{Qwen3-32B} model which has high reasoning capabilities as demonstrated in \cref{tab:overall_performance}.

Accuracy is measured as the proportion of examples for which the model outputs the correct judgment -- always %``\jpn{はい},'' 
``Yes,''
since all BIS entries are logically valid. This setup ensures that errors stem from reasoning failures, not linguistic ambiguity or semantic bias.

All models in the experiments (see \cref{model_config}) were evaluated under identical zero-shot conditions, using Japanese-language prompts with consistent formatting. No fine-tuning or task-specific adaptation was applied. Evaluation was performed on the full dataset to eliminate sampling variance and enable direct comparison of out-of-the-box reasoning robustness.

\paragraph{Model Configuration}
\label{model_config}

We evaluated prominent LLMs spanning both general-purpose and Japanese-specialized categories. The general-purpose group included OpenAI’s GPT models (\llm{GPT-5-mini}, \llm{GPT-5-nano}, \llm{GPT-4o}, \llm{GPT-4-turbo} and \llm{gpt-oss-20b}) % as well as
%Anthropic’s \llm{Claude-3-sonnet} and \llm{Claude-3-opus};
and Alibaba's \llm{Qwen3-32B}. 
These models are designed for multilingual tasks and are optimized for general reasoning performance across domains. In contrast, the Japanese-specialized models -- \llm{llm-jp-3-13b}, \llm{llm-jp-3-13b-instruct3}, \llm{llm-jp-3.1-13b-instruct4} \cite{aizawa2024llm}, and Stockmark's \llm{stockmark-13b} -- were trained or fine-tuned specifically on Japanese data, representing dedicated efforts to advance native Japanese LLM capabilities.

% Proprietary models were accessed via official APIs, with consistent configurations including temperature set to zero to ensure deterministic outputs. 
% Open-weights models were deployed locally using standardized hardware and recommended inference settings. 

All prompts were in Japanese (Basic prompt in \cref{tab:prompt_descriptions}), and all 5,000 examples in the dataset were evaluated without sampling. This setup guarantees both fairness and reproducibility across model comparisons.

\subsection{Overall Model Performance}\label{sec:overall-performance}

\cref{tab:overall_performance} summarizes the performance of each model on the \bis{} dataset, along with complementary results obtained on over 300 belief-inconsistent syllogistic reasoning samples from the NeuBAROCO benchmark. 

%\llm{GPT-4o} demonstrated the highest accuracy on both datasets, achieving 79.54\% on \bis{} and notably excelling with 94.01\% on NeuBAROCO. \llm{GPT-4-turbo} and the Japanese-specialized model \llm{llm-jp-3-13b} showed comparable results, each reaching around 59–60\% accuracy on \bis{}, but significantly lower than \llm{GPT-4o}.

The strongest performances were achieved by recent reasoning-optimized models. Both \llm{GPT-5-mini} and \llm{GPT-5-nano} with medium (default) reasoning effort achieved near-perfect accuracy on the \bis{} dataset (99.7\% and 98.8\%, respectively) while maintaining above 91\% accuracy on NeuBAROCO. Similarly, the open model \llm{Qwen3-32B} reached above 99\% accuracy with/without explicit thinking-style reasoning.

We also observed a clear trend that more recently developed models tend to exhibit stronger reasoning capabilities; for context, several deprecated early-2024 Claude models (\llm{Claude-3-sonnet-20240229} and \llm{Claude-3-opus-20240229}) were included in the table to provide a broader comparison, achieving 20.3\% and 7.2\% accuracy on the \bis{} dataset, and 78.4\% and 61.1\% on NeuBAROCO, respectively.

%\TODO{Models specifically fine-tuned on Japanese instructions, such as \llm{llm-jp-3-13b-instruct3} and \llm{stockmark-13b}, exhibited lower performance, around 40–48\%, highlighting potential shortcomings in their reasoning robustness.}
\paragraph{Performance of Japanese-specific Models}
The performance of Japanese-specialized models displayed substantial variation across generations. Earlier \llm{llm-jp} variants and \llm{stockmark-13b} lagged far behind, typically achieving between 10–60\% accuracy, which underscores their limited capacity to override belief-consistent intuitions with formal logic. However, the most recent \llm{llm-jp-3.1-13b-instruct4} exhibited a marked improvement, reaching 84.66\% accuracy—nearly on par with general-purpose reasoning-oriented models. This sharp rise suggests that newer iterations of Japanese LLMs are increasingly benefiting from fine-tuning strategies that explicitly emphasize reasoning alignment, rather than focusing solely on linguistic fluency or instruction adherence.

Notably, \llm{llm-jp-3-13b-instruct3} performed significantly below the 50\% chance level. Such results do not merely indicate random degradation of performance, but rather a systematic tendency to favor belief-consistent conclusions even when they are logically invalid. During generation, these earlier models often became ``confused,'' producing extended deliberations that acknowledge inconsistencies or factual mismatches without converging on a final decision until reaching the maximum token limit. This behavior reflects a lack of internal mechanisms for reasoning prioritization and highlights the impact of instruction-tuning objectives that insufficiently penalize belief-driven heuristics.

In contrast, the emergence of \llm{llm-jp-3.1-13b-instruct4} demonstrates that reasoning-focused refinement can substantially enhance logical consistency even within Japanese-language models. Its strong performance implies that recent Japanese LLM development has begun to integrate more explicit reasoning objectives—aligning with the broader global trend toward reasoning-centric fine-tuning observed in models such as \llm{GPT-5} and \llm{Qwen3-32B}. This transition underscores a maturing understanding that linguistic naturalness alone is insufficient; robust logical control mechanisms are essential for achieving true reasoning fidelity.

\paragraph{Reasoning Effort}
\llm{GPT-4o}, which previously dominated several reasoning benchmarks, now shows only moderate performance on \bis{} (79.5\%), though it retains high accuracy on NeuBAROCO (94.0\%). One plausible explanation is that the prompt format in our evaluation did not explicitly require models to produce elaborate reasoning chains beyond a simple ``Yes'' or ``No'' response. Since \llm{GPT-4o} does not allocate an explicit reasoning effort comparable to that of newer reasoning-enabled models, its internal deliberation may have been suppressed, resulting in lower scores on \bis{}. We will revisit this issue in more detail in \cref{sec:detailed-error-analysis-gpt4o}.

Regarding reasoning effort, a key insight emerges from the \llm{GPT-5-nano} results obtained under different reasoning-effort settings: an API parameter that controls the model’s internal resources devoted to reasoning. When this parameter was set to \textit{medium}, the model achieved 98.8\% accuracy on \bis{}, but the score dropped sharply to 69.2\% when the effort was reduced to \textit{minimum}. This finding clearly indicates that, for this dataset, substantial differences in performance arise depending on how inference-time reasoning capacity is activated and utilized.

%These results indicate that superior performance in belief-inconsistent reasoning is heavily influenced by a model's underlying training approach and inherent architectural reasoning capabilities rather than mere specialization in Japanese or model size alone. Notably, \llm{GPT-4o}'s ability to consistently achieve high accuracy suggests a strong robustness to belief bias, allowing it to distinguish logical validity effectively despite intuitive contradictions.
%
%Overall, these results make it evident that success in belief-inconsistent reasoning tasks is governed by a model’s intrinsic reasoning ability and the activation of reasoning pathways—rather than by scale, language specialization, or superficial alignment. The pronounced performance gap between different reasoning effort settings provides direct empirical evidence that the \bis{} dataset is a sensitive probe for measuring explicit reasoning competence.
%
%Overall, these results make it evident that success in belief-inconsistent reasoning tasks depends not only on the model’s intrinsic reasoning capability but also on its accessible reasoning budget. The pronounced performance gap between different \texttt{reasoning\_effort} levels provides empirical evidence that the \bis{} dataset effectively probes both the depth and robustness of reasoning activation—and exposes the extent to which belief-induced bias reemerges when reasoning computation is limited.

\paragraph{Performance Gap Between \bis{} and NeuBAROCO}
Surprisingly, Anthropic's Claude models underperformed drastically on the \bis{} dataset, with \llm{Claude-3-sonnet} and \llm{Claude-3-opus} achieving only 20.34\% and 7.18\%, respectively, despite relatively strong performances of 78.44\% and 61.07\% on NeuBAROCO.

The varying performances between datasets also highlight the sensitivity of model evaluations to task formulation and dataset characteristics. The relatively high NeuBAROCO scores of Claude models contrast sharply with their underperformance on \bis{}, emphasizing the importance of employing diverse benchmarks to comprehensively assess model reasoning capabilities, particularly in challenging belief-inconsistent contexts.

\begin{figure}[t]
\centering
\includegraphics[width=\linewidth]{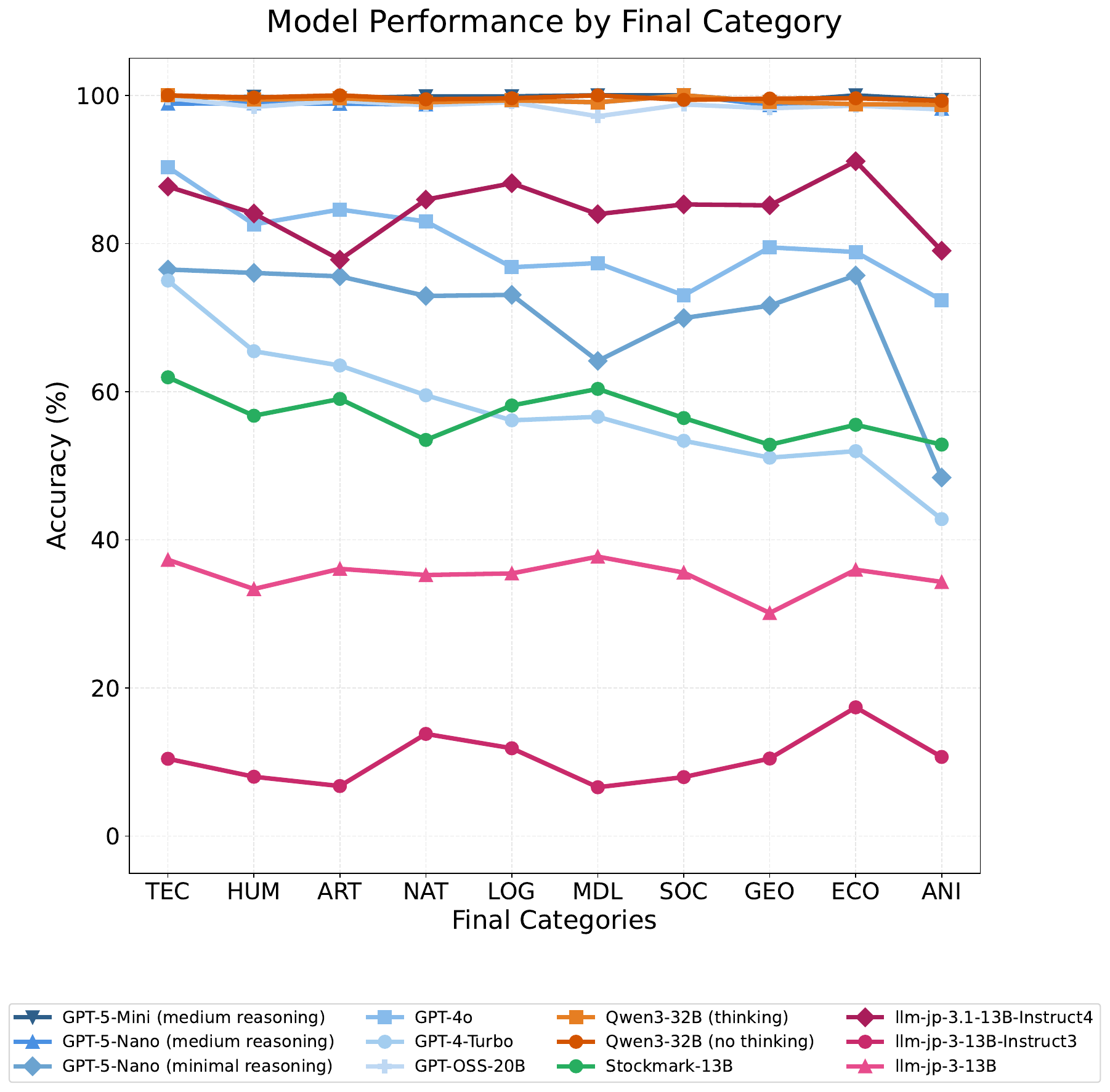}
\caption{Category-wise reasoning accuracy for each LLM and final-category.
\if0
(TEC=``Human-Made Objects \& Technology",
ART=``Music \& the Arts", 
HUM=``Human Body \& the Senses",
NAT=``Natural Phenomena \& Matter",
ECO=``Plants \& Ecosystems",
LOG=``Structures \& Logic",
MDL=``Assumptions \& Models",
SOC=``Society \& Norms",
GEO=``Topography \& Meteorology",
ANI=``Animals \& Living Things")
\fi
}
\label{table:category-wise}
\end{figure}

\begin{table*}
\small
\centering
{\scriptsize
\begin{tabular}{@{}lp{6cm}p{6cm}@{}}
\toprule
Prompt Type 
& Prompt Description (Japanese) 
& Prompt Description (English)
\\
\midrule
\vspace{5pt}
Basic 
& \jpn{以下の三段論法を考えてください：結論は前提から論理的に導き出されますか？『はい』か『いいえ』で答えてください。[...]}
&
Consider the following syllogism: Is the
conclusion logically derived from the
premises? Answer "yes" or "no." [...]
\\ 
\vspace{5pt}
Focus Logic 
&
\jpn{これは信念の不一致を含む可能性のある論理推論のサンプルです。前提と結論の論理的な関係に焦点を当て、結論が前提から確実に導き出されるかどうかを厳密に判断してください。回答は『はい』または『いいえ』のみでお願いします。[...]}
&
This is a sample of logical reasoning that may
involve belief inconsistency. Focus strictly on
the logical relationship between premises and
conclusion, and determine rigorously if the
conclusion follows necessarily from the
premises. Answer only "yes" or "no." [...]
\\
\vspace{5pt}
Chain-of-Thought
&
\jpn{以下の三段論法について、結論が前提から論理的に導き出されるかどうかを段階的に考えてください。まず、前提を分析し、次にそれらの関係性を検討し、最後に結論が論理的に妥当であるかを判断してください。思考プロセスを詳細に記述した後、最終的な回答として『はい』または『いいえ』を明確に示してください。[...]}
&
Consider the following syllogism step-by-step
to determine if the conclusion logically follows
from the premises. First, analyze the premises,
then consider their relationship, and finally
judge if the conclusion is logically valid.
Provide a detailed reasoning process and
clearly state your final answer as "yes" or
"no." [...]
\\
\vspace{5pt}
Polite 
&\jpn{以下の三段論法につきまして、結論が前提から論理的に導き出されるかどうか、ご判断いただけますでしょうか。『はい』または『いいえ』にてお答えいただけますと幸いです。[...]
}
&
Could you please determine whether the
conclusion logically follows from the premises
in the following syllogism? I would appreciate
it if you answer with either "yes" or "no." [...]
\\
Casual &
\jpn{この三段論法、どうかな？結論、前提から合ってる？『はい』か『いいえ』で教えて。[...]}
&
What do you think about this syllogism? Does
the conclusion match with the premises? Let
me know with "yes" or "no." [...]
\\ \bottomrule
\end{tabular}
}
\caption{Descriptions of prompt types.
``Basic'' prompt was used for general evaluation, and other prompts were used for re-evaluating \llm{GPT-4o} errors.}
\label{tab:prompt_descriptions}
\end{table*}

\paragraph{Category-wise Performance Analysis}

The proposed dataset also allows for assessment by category.
\cref{table:category-wise} presents the reasoning performance across individual categories.
In general, the success of belief-inconsistent reasoning does not significantly depend on the topic for most models examined.
Top-tier models (\llm{Qwen3-32B}, \llm{gpt-oss-20b} and \llm{GPT-5} models with medium reasoning effort) simply excel at counterfactual reasoning, with no notable disparities across the various categories.
In contrast, while the LLM-jp models exhibit a degree of diversity, their performance levels are generally observed to be lower.

Conversely, GPT models with lower reasoning performance (\llm{GPT-4o}, \llm{GPT-4-turbo} and \llm{GPT-5-nano} with minimal reasoning effort) demonstrate a moderate trend %relatively clear correlation 
in their strengths and weaknesses. These models perform the best in TEC, well in HUM, ART and NAT, yet struggle in prioritizing formal logic over common beliefs in ANI (Animals \& Living Things). 
Although the specific reasons remain unidentified due to the proprietary nature of the GPT models, it is plausible that such trends can be attributed to the topic distribution in the training datasets utilized by the development team.

Interestingly, we can also observe moderate similarity between \llm{llm-jp-3-13B-Instruct3} and \llm{llm-jp-3.1-13B-Instruct4}. Their relative strengths and weaknesses correlate in a comparable manner, which may indicate the presence of shared belief biases in their training corpora.

\subsection{Detailed Error Analysis for \llm{GPT-4o}}
\label{sec:detailed-error-analysis-gpt4o}
\begin{figure}[t]
\centering
\includegraphics[width=0.95\linewidth]{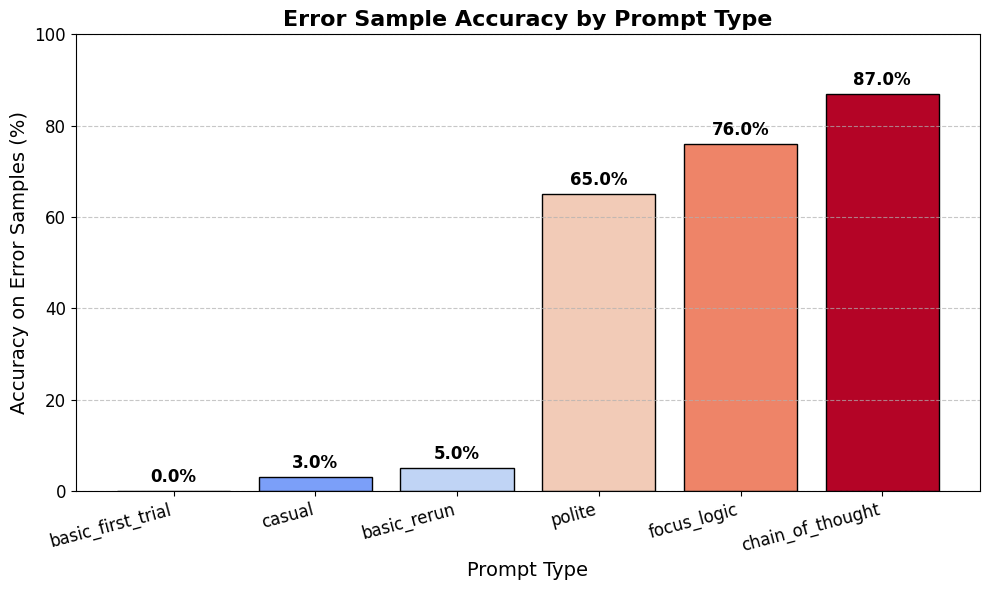}
\caption{Error sample accuracy by prompt type for \llm{GPT-4o}.}
\label{fig:error_prompt_accuracy}
%\end{figure}
%\begin{figure}
\centering
\includegraphics[width=0.95\linewidth]{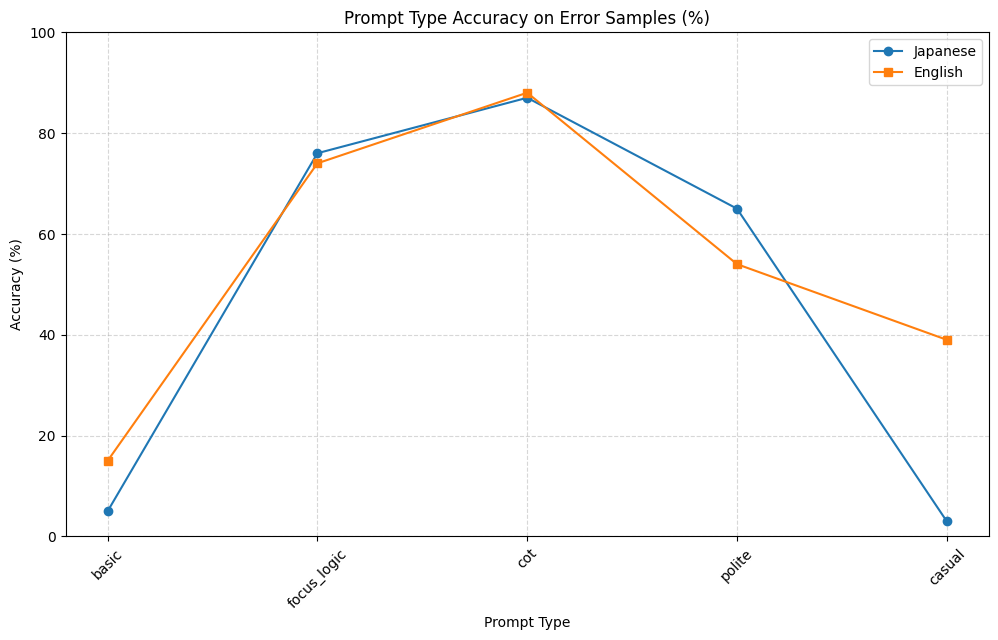}
\caption{Prompt type accuracy on error samples (\%) – retest with English prompts.}
\label{fig:english_retest}
\end{figure}

To further investigate \llm{GPT-4o}’s behavior on belief-inconsistent syllogisms, we conducted additional experiments on 100 cases where \llm{GPT-4o} initially failed. These error samples were reassessed using several carefully designed prompts, each emphasizing different reasoning approaches, linguistic styles, or explicit instructions about the belief-inconsistent nature of the task. \cref{fig:error_prompt_accuracy} summarizes the accuracy results across these varied prompts.

\paragraph{Prompt Impact Analysis} The prompt emphasizing an explicit \textit{chain-of-thought} (CoT) reasoning strategy yielded the highest accuracy improvement (87\%) among the previously failed samples. This result clearly demonstrates \llm{GPT-4o}’s latent reasoning capabilities, significantly activated when explicitly guided through structured logical steps.

Similarly, the \textit{focus\_logic} prompt, explicitly mentioning the possibility of belief inconsistency and urging strict logical evaluation, substantially improved \llm{GPT-4o}’s accuracy (76\%). This finding suggests that \llm{GPT-4o} is sensitive to explicit instructional framing and context-setting, effectively reducing its reliance on superficial plausibility heuristics when directed accordingly.

Conversely, informal prompts (\textit{casual}) and simple instruction (\textit{basic}) achieved extremely low recovery rates (3\% and 5\%, respectively). These prompts appear insufficient in addressing \llm{GPT-4o}'s belief bias, indicating that the model defaults to commonsense heuristics without clear guidance.

The polite % (\jpn{敬語}, keigo) % Commented out this because other prompts are also based on keigo
prompt yielded a moderate accuracy improvement (65\%), suggesting linguistic politeness cues may moderately encourage \llm{GPT-4o} to engage in deeper reflection or careful reasoning, though less effectively than explicitly logical or CoT framing.

Repeating this experiment with English-language prompts while keeping the Japanese data constant yielded a similar performance pattern, as illustrated in \cref{fig:english_retest}. However, the accuracy gaps were less pronounced compared to the Japanese prompts. This narrower gap in performance likely stems from \llm{GPT-4o} being extensively trained in English, enhancing its robustness to varied prompt styles. Nevertheless, this confirms the key insight: prompt design significantly influences \llm{GPT-4o}'s performance on challenging tasks like belief-inconsistent syllogisms.

\cref{tab:prompt_descriptions} provides detailed descriptions of each prompt type used in the error re-evaluation.

\paragraph{Implications for Model Deployment} These findings demonstrate the substantial impact of prompt engineering on \llm{GPT-4o}'s logical reasoning capabilities, particularly in overcoming belief bias. Although \llm{GPT-4o} already performs strongly relative to other models, explicit prompting -- such as instructing it to follow a structured reasoning process or clearly signaling the presence of belief-inconsistent content -- markedly enhances its logical consistency. Consequently, when deploying LLMs, especially in contexts demanding precise and unbiased logical inference, strategic prompt design is crucial. Clear, structured instructions, emphasizing logical rigor and explicitly guiding step-by-step reasoning, can significantly mitigate intuitive biases inherent to the model, thus ensuring more reliable and accurate outcomes.

\section{Discussion}

Our findings reveal persistent weaknesses in belief-inconsistent reasoning across LLMs. Models optimized for reasoning, such as \llm{Qwen3-32B} and \llm{GPT-5}, maintain high logical fidelity, while Japanese-specialized or alignment-heavy models often favor believable but invalid conclusions. This confirms that linguistic fluency and reasoning robustness are distinct capabilities. However, the latest \llm{llm-jp-3.1-13b-instruct4} marks a clear improvement over earlier llm-jp variants, indicating that recent Japanese models are beginning to incorporate stronger reasoning objectives. This trend aligns with the global movement toward reasoning-centric fine-tuning observed in post-2024 LLM generations.

Prompt design and inference-time reasoning effort strongly affect outcomes. Explicit logical instructions or chain-of-thought prompting significantly reduce belief bias, whereas minimal reasoning settings lead to sharp accuracy drops. Alignment-focused training, as seen in some Claude models, may further suppress acceptance of counterintuitive yet valid conclusions.

Model scale alone does not ensure logical reliability. Instead, reasoning-oriented training and architecture play a decisive role. These insights emphasize the need for benchmarks like \bis{} that expose belief bias and test logic over intuition. For reliable deployment in law, healthcare, and research, LLMs must be evaluated and optimized for strict logical consistency, not just fluency or alignment.

\section{Limitations}

While \bis{} reveals important insights into the reasoning capabilities of LLMs, this study has several limitations that should be considered. First, the evaluation focuses exclusively on syllogistic reasoning. While syllogisms offer a controlled and interpretable format, they represent only one class of logical reasoning. The results may not generalize to other reasoning forms such as causal inference, probabilistic reasoning, or multi-hop deductive chains.

Second, our evaluation relies on a single prompt design in a zero-shot setting. While this approach offers a consistent testbed, it may not fully capture the capabilities of models that perform better under more advanced prompting strategies or tailored task formulations. Prompt sensitivity remains an open variable, and performance may vary under alternative instructions or reasoning scaffolds.

Third, our scoring metric is binary and does not account for partially correct reasoning or near-misses. Models that correctly identify the logical structure but misword the conclusion, or those that reason correctly but fail to flag belief conflict, are treated the same as entirely incorrect responses.

In terms of model coverage, while we include both general-purpose and Japanese-specialized LLMs, our selection remains limited. Notably, open-weights models outside of the LLM-jp and Stockmark ecosystems, as well as mid-sized multilingual models, are not represented in this evaluation.

Lastly, model capabilities are evolving rapidly. The results presented reflect the state of model behavior at a specific point in time (early-to-mid 2025), and future model updates could yield different performance profiles.

\section{Conclusion}

We introduced \bis, the first large-scale Japanese benchmark explicitly designed to evaluate belief-inconsistent syllogistic reasoning in large language models. Our experiments show that even advanced systems still struggle when logical validity conflicts with intuitive or factual beliefs, indicating that belief bias remains a fundamental limitation across architectures and training paradigms.

Among all evaluated models, reasoning-optimized systems such as \llm{Qwen3-32B} and \llm{GPT-5} achieved near-human logical consistency, while earlier Japanese-specialized models performed significantly lower despite superior linguistic fluency. The marked improvement of \llm{llm-jp-3.1-13b-instruct4} suggests that recent Japanese LLMs are beginning to integrate reasoning-aligned fine-tuning, echoing global trends in reasoning-centric model design.

\bis{} fills a critical gap in Japanese-language reasoning evaluation by providing a natural, belief-challenging benchmark that exposes logical bias and tests reasoning fidelity under cognitive conflict. We hope this work accelerates the development of bias-resistant, logic-grounded LLMs and fosters research into reasoning alignment for safety-critical applications in law, medicine, and science.

% \section*{Acknowledgements}

% We would like to thank Yusuke Oda, Kiyomaru Hirokazu, Maki Matsuda, and Kouta Nakayama, among others, for their valuable input -- ranging from ideas for evaluating large language models, suggestions that improved the literature review, to support in drafting annotation guidelines.

\section{Bibliographical References}\label{sec:reference}

\bibliographystyle{lrec2026-natbib}
\bibliography{combine,additional_literature}

\end{document}